\pdfoutput=1
\documentclass{article} 
\usepackage[final]{colm2026_conference}

\usepackage{microtype}
\usepackage{hyperref}
\usepackage{url}
\usepackage{booktabs}
\usepackage{array}
\usepackage{graphicx}
\usepackage{multirow}
\usepackage{amsmath}
\usepackage{amssymb}
\usepackage{tikz}
\usetikzlibrary{arrows.meta,positioning,fit,backgrounds,calc}
\usepackage{enumitem}

\usepackage{lineno}

\definecolor{darkblue}{rgb}{0, 0, 0.5}
\hypersetup{colorlinks=true, citecolor=darkblue, linkcolor=darkblue, urlcolor=darkblue}

\usepackage{caption}
\title{Role-Specialized Mixture-of-Agents with Open-Weight LLMs for Clinical Prediction}

\author{
Jun Hou \\
Virginia Tech \\
\texttt{junh@vt.edu}
\And
Yi Fang \\
Virginia Tech \\
\texttt{yif@vt.edu}
\And
Xuan Wang \\
Virginia Tech \\
\texttt{xuanw@vt.edu}
}

\begin{document}

\ifcolmsubmission
\linenumbers
\fi

\maketitle

\begin{abstract}
Large Language Models (LLMs) are increasingly applied to clinical prediction tasks such as in-hospital mortality and readmission from electronic health records (EHRs). Privacy and compliance constraints motivate systems that can be deployed locally, which has increased interest in open-weight multi-agent designs. However, most medical multi-agent systems are evaluated as a single block, leaving unclear which agent role contributes to prediction and whether retrieval drives observed gains. We study a role-specialized Mixture-of-Agents (MoA) that combines medical knowledge retrieval with contrastive similar-patient reasoning. By varying the role design while holding the retrieval setup fixed, we localize the main effect to the final integrator. Pairing large open-weight analysts with a small open-weight integrator matches closed-model prompting on F1 for mortality prediction while flagging substantially more true high-risk patients. Mechanism analysis shows the role assignment directly yields a high-recall operating point without threshold tuning. The effect is task-dependent, with smaller gains for readmission because the available records correlate weakly with this longer-horizon outcome. These results position role design as a key factor in privacy-constrained, training-free clinical LLM prediction.

\end{abstract}

\section{Introduction}
\label{sec:intro}
Predicting outcomes such as in-hospital mortality and readmission from electronic health records (EHRs) is a core clinical decision-support task, and large language models (LLMs) are increasingly being adapted to this setting. Although proprietary closed-source LLMs often perform strongly under prompting, their use in clinical prediction raises privacy and compliance concerns~\citep{llmprivacy2025}. Fine-tuning a local model can address some of these concerns, but it requires labeled clinical cohorts, substantial training resources, and may limit generalization across scenarios or institutions. This creates a practical need for open-weight, locally hosted, and training-free clinical LLM systems that can approach closed-source performance. 

Multi-agent systems (MAS) offer one route toward this goal. By distributing reasoning, medical MAS frameworks allow agent roles to be specialized for distinct reasoning functions, combine retrieved evidence with task-specific deliberation, and reduce reliance on a single closed model~\citep{medagents2024, mdagents2024, colacare2025}. Yet existing systems are often evaluated as a single block, leaving unclear which role drives prediction and whether open-weight LLMs can be assembled to match closed-source models without training.

We answer these questions with a role-specialized Mixture-of-Agents built from open-weight LLMs\footnote{Code and experimental artifacts: https://github.com/JuneHou/moa-clinical-rag.git}: two contrastive analysts compare the target patient with positive- and negative-outcome exemplars, and a final integrator combines their assessments to make a  decision. We evaluate the system on the MIMIC datasets~\citep{johnson2016mimic, johnson2023mimic}, keeping knowledge sources fixed to isolate role and model effects. Our key finding is that using large open-weight analysts with a small open-weight integrator matches the closed-model few-shot baseline on MIMIC-IV mortality F1 while identifying substantially more high-risk patients, and a role-isolation analysis localizes this gain to the integrator, which shifts the decision threshold. Further analysis suggests the smaller integrator applies a lower threshold to comorbidity severity, improving mortality sensitivity but not readmission, where pre-discharge records carry less signal about post-discharge outcomes. Our contributions are threefold. \underline{First}, we introduce a training-free, open-weight role-specialized Mixture-of-Agents, deployable locally under privacy constraints. \underline{Second}, we develop a role-isolation analysis for clinical multi-agent prediction and identify the integrator as the threshold-setting role in the pipeline. \underline{Third}, we provide a mechanism analysis showing that the threshold reflects comorbidity severity that model families translate, and is bounded by where each outcome's predictive signal lies.

\section{Related Work}
\label{sec:related}
\subsection{Knowledge-augmented EHR prediction}
Knowledge augmentation has long been used to address sparsity and limited semantic structure in EHR prediction. Recent work augments clinical predictors with external medical knowledge. GRAM \citep{gram2017} incorporates ontology ancestors through attention over medical code hierarchies, KAME \citep{kame2018} uses knowledge-level attention over medical ontologies, and PRIME \citep{prime2018} injects prior medical knowledge through posterior regularization. Recent LLM-era methods move toward retrieval and generated medical knowledge. GraphCare \citep{graphcare2024} builds patient-specific knowledge graphs
from LLM-extracted and biomedical-graph evidence, RAM-EHR \citep{ramehr2024} retrieves textual medical
knowledge for clinical prediction, and KARE \citep{kare2025} combines knowledge-graph community
retrieval with LLM reasoning and fine-tuning. These methods improve prediction but require
task-specific training, fine-tuning, or custom knowledge construction.

KARE is the most closely related method to ours, reached state-of-the-art EHR prediction by distilling closed-source LLM reasoning into a small language model. Rather than focusing on improving a trained predictor, we study a training-free and open-weight role design, and investigate how different agent roles affect the behavior of reasoning and clinical prediction.

\subsection{Multi-agent medical reasoning and role specialization}
Multi-agent methods have shown enhanced reasoning through iterative debate \citep{mad2023, reconcile2024} or by aggregating independent
outputs, as in Mixture-of-Agents \citep{moa2024}. In medical domain,
MedAgents \citep{medagents2024} and MDAgents \citep{mdagents2024} use role-playing or adaptive
collaboration for medical reasoning, and ColaCare \citep{colacare2025} brings 
collaboration to EHR modelling. These systems motivate collaborative reasoning for medical tasks, and some include useful component analyses such as varying agent number, collaboration structure, retrieval support, or adaptive complexity assignment. However, they do not isolate, in an EHR prediction setting, which agent role controls the operating point at which patients are flagged.

We study role assignment as the primary experimental variable within a fixed Mixture-of-Agents workflow. Specialized analysts each produce a contrastive reading against a different historical exemplar, while the integrator reconciles their outputs into the final probabilistic prediction.

\section{Methods}

\begin{figure}[t]
\centering
\includegraphics[width=\textwidth]{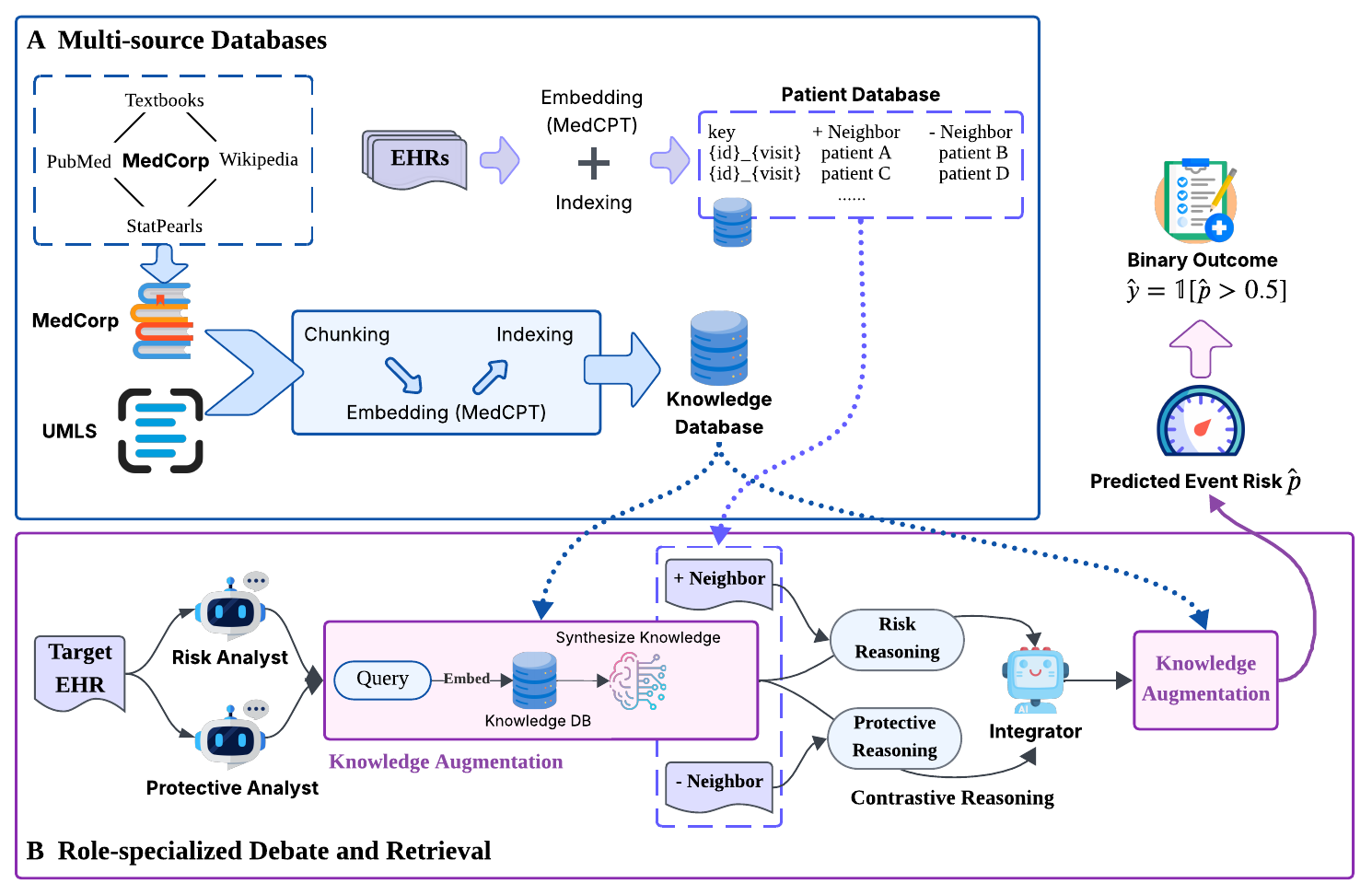}
\caption{Pipeline overview. \textbf{(A) Multi-source databases.} A knowledge database is built from MedCorp and UMLS, and a patient database is built from the EHRs. \textbf{(B) Role-specialized retrieval and integration.} The label-blind risk analyst and protective analyst read the target record and augment their reasoning through retrieval from the knowledge database and the patient database; the integrator runs its own retrieval and integrates the two readings into a final prediction.}
\label{fig:pipeline}
\end{figure}

\label{sec:method}
Our pipeline, shown in Figure~\ref{fig:pipeline}, reaches an outcome prediction through a multi-step reasoning process distributed across three specialized roles.

\paragraph{Knowledge Database and Retrieval.} The agents use two retrieval sources, medical knowledge and patient exemplars. The medical-knowledge retriever returns evidence relevant to the target record or to a focused query from the integrator. The patient retriever returns one positive-outcome exemplar and one negative-outcome exemplar from a labeled reference pool. These two exemplars define the contrastive inputs to the analyst roles.

\paragraph{Roles in the Mixture-of-Agents.} The pipeline has three roles, two label-blind analysts and a final integrator (Figure~\ref{fig:pipeline}B). The risk analyst reads the target record together with the positive-outcome exemplar, while the protective analyst reads the target record together with the negative-outcome exemplar. Each analyst performs a contrastive comparison without label and extracts evidence relevant to the final decision. The analysts are instructed to not make an early prediction before integration. The integrator then reconciles the two analysts' reasonings, issues a focused query for additional medical knowledge, and produces the final risk score. 

Given a target patient EHR $x$, let $e^{+}$ and $e^{-}$ denote the retrieved positive- and negative-outcome exemplars. Let $k_x$ denote the medical knowledge returned by MedRAG from the target record and shared with both analysts, and let $k$ denote the knowledge returned from the integrator's focused query. The two analyst readings are denoted by $a^{+}$ and $a^{-}$. The integrator query by $q$, the predicted positive-class risk score by $\hat{p}$, and the binary prediction by $\hat{y}$. The pipeline is written as
\begin{gather*}
k_x = \mathrm{MedRAG}(x), \\
a^{+} = \mathrm{Analyst}_{\mathrm{risk}}(x, e^{+}, k_x), \qquad a^{-} = \mathrm{Analyst}_{\mathrm{prot}}(x, e^{-}, k_x), \\
q = \mathrm{Integrator}_{\mathrm{query}}(x, a^{+}, a^{-}), \qquad k = \mathrm{MedRAG}(q), \\
\hat{p} = \mathrm{Integrator}(x, a^{+}, a^{-}, k), \qquad \hat{y} = \mathbb{1}[\hat{p} > 0.5].
\end{gather*}
Rather than output a binary label directly, the integrator produces a complementary probability pair that sums to one, a positive-class probability $\hat{p}$ and a negative-class probability $1-\hat{p}$, which conveys the agent's certainty toward each outcome. We treat $\hat{p}$ as the risk score and threshold it at $0.5$ to obtain $\hat{y}$.

\section{Experiments \& Results}
\label{sec:expresults}
\subsection{Experimental setup}
\label{sec:setup}
We use KARE's train and test splits. The main results (Table~\ref{tab:main}) use the full test sets; in the column order of Table~\ref{tab:main} (MIMIC-III/IV mortality, MIMIC-III/IV readmission), these contain $n = 986/996/1013/996$ patients with positive rates of $19.16/5.42/46.50/54.82\%$. The preliminary role-isolation factorial (Table~\ref{tab:factorial}) uses the same $n{=}100$ per-cell subsample for every row. The all-large and mixed endpoint configurations are additionally evaluated on the full test sets in Table~\ref{tab:main}. Readmission is labeled positive when the patient is readmitted within 15 days of discharge, and negative otherwise. Throughout, gpt-oss-120B \citep{gptoss2025} represents the large model and Qwen2.5-7B \citep{qwen2024} represents the small model, both served locally. Our reported system pairs large analysts with a small integrator. We report Accuracy, F1 (macro-averaged), Sensitivity, and Specificity, and additionally AUROC, the area under the receiver operating characteristic curve, which measures ranking discrimination independent of the decision threshold.
\paragraph{Retrieval control.} We use the same retrieval resources, retrievers, and exemplar pools across all configurations, so only role topology and model assignment vary. Patient exemplars are derived following KARE \citep{kare2025} and kept fixed across runs. Each target patient is matched to the nearest positive-outcome exemplar and the nearest negative-outcome exemplar from the labeled reference pool using MedCPT embeddings \citep{medcpt2023} and cosine similarity. For medical knowledge retrieval, we replace KARE's custom knowledge graph with a vector database built from publicly available MedRAG/MedCorp \citep{medrag2024} and UMLS \citep{umls2004}, using MedCPT as the encoder and cosine similarity for retrieval. This avoids rebuilding a per-cohort knowledge graph for each new deployment, a process that in KARE relies on closed-weight LLM and embedding APIs. All agents decode with temperature $0.7$, and prompt templates are identical within each task and retrieval condition (Appendix~\ref{app:prompts}). We name configurations by the inputs each agent receives. \emph{CoT} denotes a single agent with neither retrieval nor exemplars. \emph{single$+$RAG$+$Sim} denotes a single agent given both. \emph{single$+$RAG} denotes the control that withholds the exemplars. \emph{multi$+$RAG$+$Sim} denotes the three-role pipeline, and \emph{multi$+$Sim} the same pipeline with the medical-knowledge retrieval removed. Where a multi-agent row pairs large analysts with a small integrator, the integrator model is given in parentheses. These names are used consistently in every table and figure.

\subsection{Preliminary results}
\label{sec:prelim}
\paragraph{The integrator is the deciding factor.} To choose the configuration used throughout the paper, we begin with a controlled role-isolation experiment that varies one role at a time.
Replacing only the integrator with the small model raises MIMIC-IV mortality Sensitivity from $13.0$ to $64.8$ and gives the best macro-F1 and within-cell AUROC on both mortality cells. Therefore, we adopt the large-analyst, small-integrator configuration as our default in the main results. Replacing only an analyst or only the retrieval model does not produce this gain, and the effect is invariant to model family, reproducing when a second model fills the analyst and retrieval roles. Because the integrator is the only role whose replacement substantially shifts performance, it is the role that sets the decision threshold. On readmission there is no stable best role and model pair, so its readmission rows are honest negatives, with details in Appendix~\ref{app:factorial-all}.

\begin{table}[t]
\centering\small
\setlength{\tabcolsep}{4pt}
\caption{Role-isolation factorial, both mortality cells. A/R/I $=$
analyst/retrieval/integrator; oss $=$ GPT-OSS-120B, qwen $=$ Qwen2.5-7B; oss CoT $=$
chain-of-thought. F1/Sens/Spec in \%; AUROC computed within each cell. \textbf{Bold} (large analysts,
small integrator) best on both.}
\label{tab:factorial}
\begin{tabular}{l cccc cccc}
\toprule
& \multicolumn{4}{c}{\textbf{MIMIC-III Mort.}} & \multicolumn{4}{c}{\textbf{MIMIC-IV Mort.}}\\
\cmidrule(lr){2-5}\cmidrule(lr){6-9}
A/R/I & MacF1 & Sens & Spec & AUROC & MacF1 & Sens & Spec & AUROC \\
\midrule
oss/oss/oss   & 40.0 & 25.9 & 58.7 & .406 & 38.5 & 13.0 & 80.4 & .543 \\
oss/qwen/oss  & 38.3 & 18.5 & 67.4 & .399 & 35.0 & 7.4  & \textbf{84.8} & .533 \\
qwen/oss/oss  & 36.2 & 24.1 & 52.2 & .340 & 36.3 & 11.1 & 78.3 & .517 \\
\textbf{oss/oss/qwen} & \textbf{47.3} & \textbf{75.9} & 23.9 & \textbf{.470} & \textbf{57.4} & \textbf{64.8} & 50.0 & \textbf{.603} \\
oss/qwen/qwen & 41.3 & 24.1 & 65.2 & .444 & 40.6 & 18.5 & 73.9 & .575 \\
qwen/oss/qwen & 40.3 & 27.8 & 56.5 & .420 & 49.0 & 33.3 & 69.6 & .544 \\
qwen/qwen/qwen & 41.4 & 72.2 & 17.4 & .468 & 54.6 & 59.3 & 50.0 & .491 \\
oss CoT  & 45.9 & 46.3 & 45.7 & .454 & 40.9 & 22.2 & 67.4 & .536 \\
\bottomrule
\end{tabular}
\end{table}

\subsection{Main results}
\label{sec:results}
The results are shown in Table~\ref{tab:main} for the mortality prediction and readmission prediction tasks on MIMIC-III and MIMIC-IV datasets. On MIMIC-IV mortality, our role-specialized pipeline matches API-based Claude-3.5 few-shot prompting on F1 ($+1.4$ points) while improving Sensitivity by $45.3$ points, and improves on the zero-shot setting by $11.8$ points in F1 and $58.4$ points in Sensitivity.
 Compared with task-supervised pre-trained baselines on MIMIC-IV mortality, our pipeline has lower F1 than GRU, AdaCare, and GraphCare by $17.6$--$21.5$ points, but higher Sensitivity by $3.3$--$18.2$ points. The same pattern appears on MIMIC-III mortality, and readmission shows the same shifting in recall but with weaker F1. Across the open models, the small model predicts positive more often and gives high Sensitivity, while the large model predicts negative more often and gives high Accuracy and F1 on the majority label. The all-negative majority class baseline shows that high Accuracy on imbalanced mortality datasets can come from predicting all patients as negative. Our pipeline instead achieves high nonzero Sensitivity, indicating that it identifies positive cases rather than relying on majority class guessing.
On the full MIMIC-IV mortality test set, AUROC is $0.763$ for the single large model, $0.730$ for the all-large pipeline, and $0.706$ for the mixed assignment. The single large model therefore retains stronger ranking performance, while the mixed assignment reaches higher Sensitivity at the default cutoff. Full AUROC and AUPRC results are reported in Appendix~\ref{app:full}.


\begin{table}[t]
\centering\small
\setlength{\tabcolsep}{4pt}
\caption{Main results, all four cells (Acc/F1/Sens, \%), by training regime. Our training-free rows use RAG $+$ KARE exemplars (single vs.\ multi $=$ topology).
\textbf{Bold} marks the best training-free row. In the open rows oss $=$ GPT-OSS-120B and qwen $=$ Qwen2.5-7B; multi rows are named by their (analyst/retrieval/integrator) assignment. }
\label{tab:main}
\begin{tabular}{l ccc ccc ccc ccc}
\toprule
& \multicolumn{3}{c}{\textbf{MIMIC-IV Mort.}} & \multicolumn{3}{c}{\textbf{MIMIC-III Mort.}}
& \multicolumn{3}{c}{MIMIC-IV Read.} & \multicolumn{3}{c}{MIMIC-III Read.}\\
& \multicolumn{3}{c}{\scriptsize(pos 19.2\%)} & \multicolumn{3}{c}{\scriptsize(pos 5.4\%)}
& \multicolumn{3}{c}{\scriptsize(pos 46.5\%)} & \multicolumn{3}{c}{\scriptsize(pos 54.8\%)}\\
\cmidrule(lr){2-4}\cmidrule(lr){5-7}\cmidrule(lr){8-10}\cmidrule(lr){11-13}
Method & Acc & F1 & Sens & Acc & F1 & Sens & Acc & F1 & Sens & Acc & F1 & Sens \\
\midrule
All-negative & 80.8 & 44.7 & 0.0 & 94.6 & 48.6 & 0.0 & 53.5 & 34.9 & 0.0 & 45.2 & 31.1 & 0.0 \\
\midrule
\multicolumn{13}{l}{\emph{Pre-trained (task-supervised, trained from scratch)}}\\
GRU            & 88.7 & 76.4 & 42.9 & 92.7 & 50.7 & 3.7  & 62.4 & 62.2 & 68.3 & 62.2 & 61.5 & 68.9 \\
AdaCare        & 88.7 & 78.2 & 50.3 & 90.6 & 54.1 & 9.1  & 62.9 & 62.9 & 58.4 & 61.6 & 60.5 & 70.8 \\
GraphCare      & 91.5 & 80.3 & 57.8 & 94.9 & 58.3 & 17.2 & 65.7 & 65.5 & 66.2 & 65.4 & 64.1 & 70.3 \\
\midrule
\multicolumn{13}{l}{\emph{Fine-tuned LLM (ceiling)}}\\
KARE           & 94.1 & 90.4 & 73.2 & 95.3 & 64.6 & 24.7 & 73.9 & 73.8 & 85.6 & 73.9 & 73.7 & 76.7 \\
\midrule
\multicolumn{13}{l}{\emph{Instruction-tuned, training-free (closed)}}\\
Claude 0-shot  & 80.5 & 47.0 & 2.7  & 89.5 & 50.4 & 6.4  & 49.4 & 45.7 & 81.8 & 54.3 & 35.4 & 98.9 \\
Claude few-shot& 84.5 & 57.4 & 15.8 & 91.5 & 53.5 & 13.7 & 54.1 & 51.9 & 75.2 & 57.1 & 49.3 & 75.5 \\
\cmidrule(lr){1-13}
\multicolumn{13}{l}{\emph{Instruction-tuned, training-free (open, ours)}}\\
oss single     & 71.6 & 52.8 & 22.1 & 61.2 & 42.0 & 33.3 & 47.7 & 47.7 & 48.8 & 43.8 & 43.1 & 49.6 \\
oss/oss/oss    & 78.7 & 54.4 & 14.7 & 74.4 & 46.9 & 22.2 & 47.5 & 47.2 & 43.5 & 48.8 & 48.8 & 42.1 \\
qwen single    & 37.4 & 37.4 & 94.7 & 11.4 & 11.4 & 98.1 & 46.8 & 34.3 & 97.2 & 54.6 & 35.5 & 99.5 \\
qwen/qwen/qwen & 65.6 & 58.0 & 60.0 & 40.0 & 33.0 & 72.2 & 49.9 & 41.0 & 95.3 & 54.8 & 39.4 & 96.0 \\
\textbf{oss/oss/qwen} & 66.4 & \textbf{58.8} & \textbf{61.1} & 36.4 & 30.2 & 59.3 & 46.1 & 34.7 & 94.5 & 53.4 & 36.2 & 96.2 \\
\bottomrule
\end{tabular}
\end{table}

\section{Model Analysis}
\label{sec:ablation}
The main results show that the role-specialized pipeline reaches higher Sensitivity than single agent and closed prompting at competitive F1. Having fixed the default to large analysts with a small integrator in the preliminary analysis, we now study where this behavior comes from. We vary model placement while keeping the role instructions fixed, so our analysis concerns model assignment rather than role semantics. We first ask how the model size assigned to each role shapes the prediction, then which component carries the gain (\S\ref{sec:modules}) and how the pipeline reshapes the predicted-risk distribution (\S\ref{sec:slope}).

\paragraph{Model size in the pipeline.} When all roles use the large model, the system remains conservative, with high Specificity and low Sensitivity. When all roles use the small model, the system becomes more sensitive but loses discrimination, reaching AUROC $0.491$ on MIMIC-IV mortality. The mixed assignment, with large models as analysts and a small model as integrator, avoids both extremes. It gives the best F1, Sensitivity, and within-cell AUROC on both mortality cells in Table~\ref{tab:factorial}, including AUROC $0.603$ on MIMIC-IV mortality. These results localize the change in sensitivity to the integrator and show that model placement changes the operating point.

\subsection{Topology, retrieval, and exemplars}
\label{sec:modules}
We vary the three components of the system, the multi-agent topology, the medical-knowledge retrieval, and the similar-patient exemplars, across mortality and readmission tasks on MIMIC-III and MIMIC-IV (Table~\ref{tab:ablate}). The findings below are drawn mainly from the mortality tasks with clearest patterns.
A single agent is limited by how much context it can use well. Adding retrieved medical knowledge to a single agent lowers MIMIC-IV mortality F1, from $58.3$ for the large model's \emph{CoT} baseline to $37.7$ for \emph{single$+$RAG}, with the same direction for the small model, and it raises predicted risk on the survivors.
Adding the retrieved patient exemplars on top of medical knowledge improves the single-agent setting, from \emph{single$+$RAG} to \emph{single$+$RAG$+$Sim} (large $37.7\to52.8$, small $32.9\to37.4$), but stays below the \emph{multi$+$RAG$+$Sim} configurations on the same task, with the small single agent reaching Sensitivity $94.7$ at Specificity $23.7$, a near-all-positive predictor. The exemplar pair is thus not sufficient on its own. It becomes useful once the two neighbours are split into two contrastive analyst roles rather than placed into one long single-agent prompt.

Medical-knowledge retrieval, by contrast, is not the main source of the gain. Removing MedRAG from \emph{multi$+$RAG$+$Sim} has limited effect on F1 For the pipeline with all large models, F1 changes from $54.4$ for \emph{multi$+$RAG$+$Sim} to $51.3$ for \emph{multi$+$Sim}. The multi-agent topology instead makes the system more robust to retrieved context by distributing how that context is read and integrated, so the benefit of retrieval depends not only on what is retrieved but also on how retrieval and reasoning are organized across the system.

Overall, the multi-agent topology is the component that carries the gain, the contrastive patient exemplars help only once they are split across the two analyst roles, and the medical-knowledge retrieval is close to neutral.

\begin{table}[t]
\centering\small
\setlength{\tabcolsep}{4.5pt}
\caption{Component ablation, all four cells (\%). \emph{Topo} single vs.\ multi; \emph{RAG} MedRAG;
\emph{Sim} KARE exemplars; the \emph{single$+$RAG} no-exemplar row is the added control. The rows reported in the
main results table are the \emph{single$+$RAG$+$Sim} and \emph{multi$+$RAG$+$Sim} configurations.}
\label{tab:ablate}
\resizebox{\textwidth}{!}{%
\begin{tabular}{l ccc cccc cccc cccc cccc}
\toprule
& & & & \multicolumn{4}{c}{\textbf{MIMIC-IV Mort.}} & \multicolumn{4}{c}{MIMIC-III Mort.}
& \multicolumn{4}{c}{MIMIC-IV Read.} & \multicolumn{4}{c}{MIMIC-III Read.}\\
\cmidrule(lr){5-8}\cmidrule(lr){9-12}\cmidrule(lr){13-16}\cmidrule(lr){17-20}
Model & Topo & RAG & Sim & Acc & F1 & Sens & Spec & Acc & F1 & Sens & Spec & Acc & F1 & Sens & Spec & Acc & F1 & Sens & Spec \\
\midrule
\multirow{5}{*}{GPT-oss}
 & single & --  & --  & 77.9 & 58.3 & 24.2 & 90.7 & 70.6 & 48.4 & 46.3 & 72.0 & 49.0 & 45.2 & 24.6 & 70.1 & 43.3 & 35.0 & 7.0  & 87.3 \\
 & single & \checkmark & \checkmark & 71.6 & 52.8 & 22.1 & 83.4 & 61.2 & 42.0 & 33.3 & 62.8 & 47.7 & 47.7 & 48.8 & 46.7 & 43.8 & 43.1 & 49.6 & 36.7 \\
 & single & \checkmark & --  & 48.5 & 37.7 & 17.9 & 55.8 & 59.7 & 42.0 & 40.7 & 60.8 & 49.1 & 49.0 & 56.5 & 42.6 & 51.9 & 46.6 & 76.2 & 22.4 \\
 & multi  & --  & \checkmark & 78.3 & 51.3 & 10.0 & 94.6 & 80.0 & 50.5 & 25.9 & 83.1 & 50.1 & 48.9 & 37.4 & 61.3 & 47.5 & 46.9 & 33.9 & 64.0 \\
 & multi  & \checkmark & \checkmark & 78.7 & 54.4 & 14.7 & 94.0 & 74.4 & 46.9 & 22.2 & 77.4 & 47.5 & 47.2 & 43.5 & 50.9 & 48.8 & 48.8 & 42.1 & 56.9 \\
\midrule
\multirow{5}{*}{Qwen}
 & single & --  & --  & 68.6 & 57.9 & 47.4 & 73.7 & 50.6 & 39.7 & 74.1 & 49.3 & 48.6 & 42.7 & 86.6 & 15.5 & 53.3 & 37.5 & 94.5 & 3.3  \\
 & single & \checkmark & \checkmark & 37.4 & 37.4 & 94.7 & 23.7 & 11.4 & 11.4 & 98.1 & 6.5  & 46.8 & 34.3 & 97.2 & 3.0  & 54.6 & 35.5 & 99.5 & 0.2  \\
 & single & \checkmark & --  & 33.0 & 32.9 & 96.8 & 17.8 & 10.5 & 10.5 &100.0 & 5.4  & 47.0 & 36.0 & 95.1 & 5.2  & 54.9 & 36.4 & 99.3 & 1.1  \\
 & multi  & --  & \checkmark & 68.1 & 58.8 & 53.7 & 71.5 & 42.6 & 34.2 & 63.0 & 41.4 & 48.2 & 40.3 & 90.9 & 11.1 & 53.5 & 40.6 & 91.4 & 7.6  \\
 & multi  & \checkmark & \checkmark & 65.6 & 58.0 & 60.0 & 66.9 & 40.0 & 33.0 & 72.2 & 38.1 & 49.9 & 41.0 & 95.3 & 10.3 & 54.8 & 39.4 & 96.0 & 4.9  \\
\bottomrule
\end{tabular}
}
\end{table}

\subsection{Probability-level separation}
\label{sec:slope}

\begin{figure}[t]
\centering
\includegraphics[width=\textwidth]{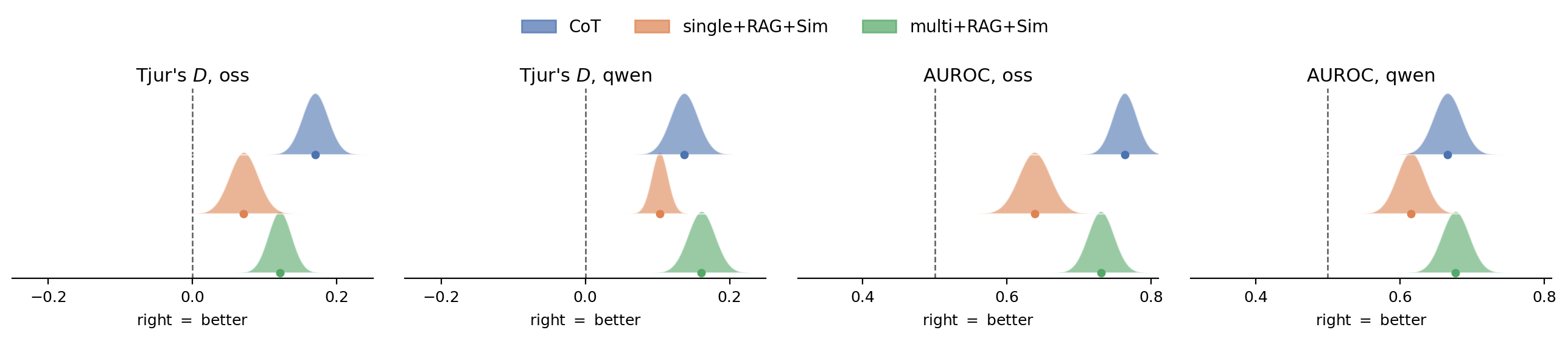}
\caption{Tjur's $D$ (left two panels) and AUROC (right two) on MIMIC-IV mortality, for \emph{CoT}, \emph{single$+$RAG$+$Sim}, and the \emph{multi$+$RAG$+$Sim} pipeline. Each curve is the $95\%$ bootstrap CI for that condition and the dot is its point estimate, with the dashed line marking the null ($D{=}0$ or AUROC${=}0.5$).}
\label{fig:slope-mort}
\end{figure}

\paragraph{The pipeline restores probability-level separation.} We next examine whether the component changes alter only the binary predictions or also the predicted-risk distribution. We measure this with the discrimination slope, Tjur's coefficient of discrimination, computed as the mean predicted risk on the actual positive cases minus the mean predicted risk on the actual negative cases, so a larger value means the two outcome groups are better separated (Figure~\ref{fig:slope-mort}). The full version over all four task and dataset combinations is in Appendix~\ref{sec:mech}. \emph{single$+$RAG$+$Sim} flattens this separation by raising predicted risk on the survivors, whereas \emph{multi$+$RAG$+$Sim} restores part of it. AUROC moves in the same direction, indicating that the recovery is not only a calibration artifact. The no-retrieval \emph{CoT} baseline retains the highest AUROC on this cell. The observed restoration therefore applies relative to the single agent given the same retrieved context and reflects a change in operating point. Together with the component ablation, this suggests that the pipeline does more than add retrieved context. It changes how retrieved and exemplar evidence is transformed into a risk score before the final threshold is applied.

\section{Error Analysis}
\label{sec:signal}
The sensitivity gain is localized to the integrator, and it appears on mortality but not on readmission. We analyze why the same integrator change helps one task and fails on the other. Replacing the gpt-oss integrator with the Qwen makes the system predict positive for far more patients on both tasks. On MIMIC-IV mortality, the predicted-positive rate rises by $30.2$ percentage points when the integrator changes from gpt-oss to Qwen. On readmission, the predicted-positive rate rises by $48.8$ percentage points. We trace this behavior to how each model family turns the number of recorded conditions into predicted risk. On mortality, both families read similar evidence, but they require different numbers of recorded conditions before crossing the positive threshold. On readmission, the same relationship between recorded conditions and predicted risk is applied to an outcome for which the input record contains weak code-level signal.
Within the all-large pipeline, predicted probability correlates with combined analyst's response length on mortality ($\rho=+0.33$ and $+0.56$ on MIMIC-III and MIMIC-IV), but not on readmission, where $\rho=-0.02$ and $+0.07$ (Appendix~\ref{app:taskasym})

\begin{figure}[t]
\centering
\includegraphics[width=0.95\textwidth]{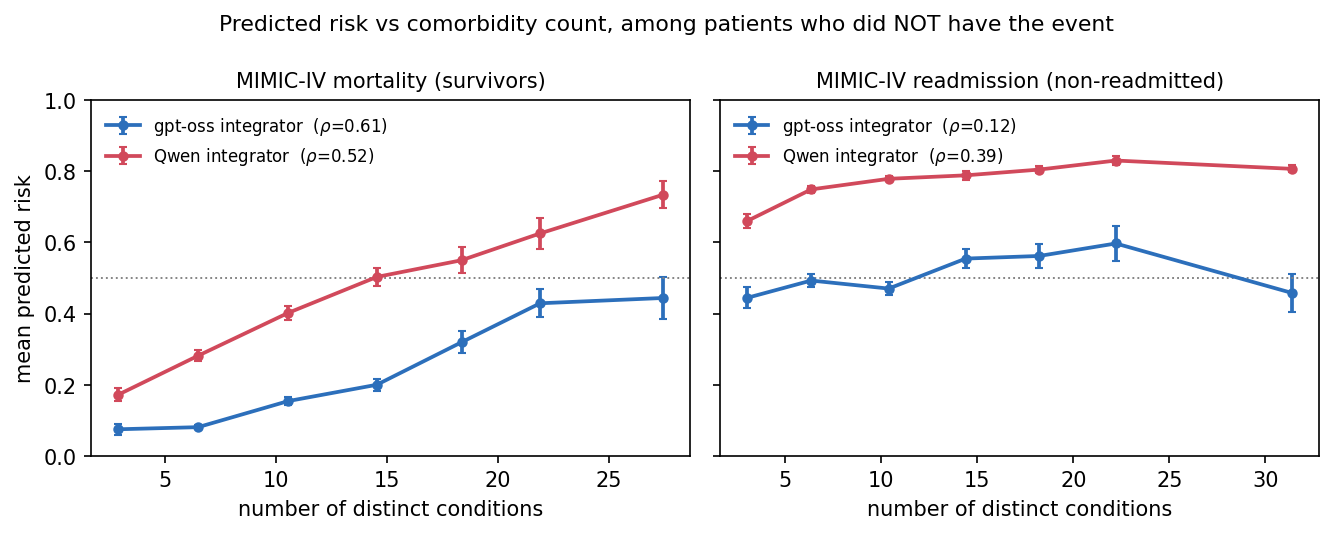}
\caption{Mean predicted risk versus number of recorded conditions, among patients who did not
have the event. Bars are standard error.}
\label{fig:dxcount}
\end{figure}

\paragraph{Comorbidity as risk, at different thresholds.} Among
patients whose event is absent, predicted risk still rises with the number of recorded conditions for both integrators (Figure~\ref{fig:dxcount}, left). The two families differ in where they place the decision threshold along this scale. The gpt-oss integrator assigns a positive prediction only when the patient has many recorded conditions and leaves many moderate-count patients negative. The Qwen integrator crosses the threshold at a lower condition count. The confusion-matrix breakdown (Appendix~\ref{app:family}) shows that both families order patients similarly by the number of recorded conditions, and that the main difference is the threshold at which this count becomes a positive prediction. The error pattern follows from this difference. Qwen turns moderate severity survivors into false positives, whereas gpt-oss leaves moderate-count decedents below threshold and creates false negatives. On MIMIC-IV mortality, $56\%$ of the decedents gpt-oss misses are patients the Qwen
integrator would flag.

\paragraph{Readmission fails with weak signal in records.} The same use of recorded conditions fails for 15-day readmission because the readmission label is weakly represented in the pre-discharge record. We support this with two measurements.
First, the signal is weak in the diagnoses, procedures, and medications that the model receives. A five-fold cross-validated bag-of-codes classifier using logistic regression and gradient boosting reaches AUROC $0.77$ on MIMIC-IV mortality but only $0.48$ to $0.59$ on readmission (Table~\ref{tab:taskasym}).The mortality and readmission settings use byte-identical inputs and differ only in the label, so the gap indicates that readmission determinants are not well captured by the available record. They likely depend on post-discharge factors such as adherence, follow-up, and social support. Second, the agents extract even less signal than this code-level reference. The unaugmented LLM reaches AUROC $0.43$ on MIMIC-IV readmission, below the $0.59$
bag-of-codes bag-of-codes result. The Qwen integrator also continues to raise predicted risk with condition count among patients who were not readmitted (Figure~\ref{fig:dxcount}, right, $\rho=0.39$), while gpt-oss is much flatter ($\rho=0.12$). Because recorded conditions separate mortality cases better than readmission cases, the Qwen integrator's lower threshold produces high readmission recall without useful separation. Sensitivity rises from $43.5$ to $94.5$, but F1 drops from $47.2$ to $34.7$ and specificity falls near $4\%$. The failure is not that the multi-agent design cannot use roles. It is that the integrator applies a mortality-like relationship between recorded conditions and predicted risk to an outcome whose signal is mostly outside the available record. Full probes are in Appendix~\ref{app:taskasym}.

\begin{table}[t]
\centering\small
\caption{Signal-location probes, mortality vs.\ readmission. \emph{Feature AUROC}, cross-validated
bag-of-codes classifier. \emph{CoT-LLM AUROC}, the unaugmented gpt-oss agent (qwen
$.625/.666/.498/.542$). \emph{Similars $\Delta$AUROC}, gpt-oss.}
\label{tab:taskasym}
\begin{tabular}{lcccc}
\toprule
Cell & Prev.\ \% & Feature AUROC & CoT-LLM AUROC & Similars $\Delta$AUROC \\
\midrule
mortality m4   & 19.2 & 0.77 & 0.763 & $+0.225$ \\
mortality m3   & 5.4  & 0.52 & 0.635 & $+0.009$ \\
readmission m4 & 46.5 & 0.59 & 0.431 & $-0.029$ \\
readmission m3 & 54.8 & 0.48 & 0.422 & $-0.090$ \\
\bottomrule
\end{tabular}
\end{table}

\section{Conclusion}
\label{sec:conclusion}
We study a training-free, open-weight clinical prediction pipeline where large models act as contrastive analysts and a small model integrates their outputs. On MIMIC-IV mortality, this role-specialized design matches closed-model few-shot prompting on F1 while achieving higher Sensitivity than both closed-model prompting and task-supervised pre-trained baselines. The gain stems from the integrator’s decision threshold and from differences in how model families map recorded conditions to risk. Although fine-tuned models such as KARE remain strongest overall, our results show that model placement across roles is a key factor in open-weight clinical prediction, especially in high-recall screening settings that avoid closed APIs and task-specific fine-tuning. More broadly, decomposing a clinical prediction across specialized open-weight agents is a workable design, and the most effective model family and size depend on the role and the task.

\section{Limitations}
\label{sec:limitations}
Our evaluation is limited in scope. For the mixed assignment, only MIMIC-IV mortality provides clear discrimination. All results use MIMIC and lack external or prospective validation. We also evaluate only two model families, Qwen and gpt-oss, so the behavior concluded from model family may not extend to other families. Finally, our advantage is a decision-threshold difference rather than a uniform gain. It appears as higher Sensitivity, while the pre-trained and fine-tuned baselines lead on F1 and Accuracy and we match neither the fine-tuned SOTA (KARE) nor the large model's AUROC. We therefore scope our claims to the training-free regime and to this risk-sensitive threshold. 
Each patient receives one exemplar from each outcome class. Top $k$ retrieval was not evaluated. The findings cover one analyst and integrator decomposition. There was no clinician review of intermediate outputs and no evaluation of additional outcomes such as length of hospital stay.

\section*{Acknowledgement}
This research is sponsored by NSF 2442253, 2607580, NIH 1R21AG091260-01, USDA NIFA, Commonwealth Cyber Initiative, and generous gifts from Nvidia, Cisco, and the Amazon-Virginia Tech Initiative. This research used the Delta system at the National Center for Supercomputing Applications [award OAC 2005572] through allocation [NAIRR240202] from the Advanced Cyberinfrastructure Coordination Ecosystem: Services \& Support (ACCESS) program, which is supported by National Science Foundation grants \#2138259, \#2138286, \#2138307, \#2137603, and \#2138296.

\section*{Ethics Statement}
All datasets used in this research are publicly available for research use, under the terms and licenses specified by their creators. We use MIMIC-III v1.4 and MIMIC-IV v2.0, both released under the PhysioNet Credentialed Health Data Use Agreement, which requires credentialed access and human-subjects research training. All models in our pipeline are open-weight and run locally, so no patient data is transmitted to external or third-party language-model services. No proprietary data or restricted-access resources were used. We do not display raw excerpts from any dataset in this paper. We do not attempt to identify or deanonymize patients in the data in any way during our research.

\paragraph{AI Assistance.} We used AI assistants for parts of the implementation and manuscript preparation, including generating LaTeX code for tables and refining text written by the authors. All AI-generated content was carefully reviewed and revised by the authors to ensure accuracy and clarity.

\bibliography{colm2026_conference}
\bibliographystyle{colm2026_conference}

\appendix
\section{AUROC / AUPRC (score-based metrics)}
\label{app:full}
Table~\ref{tab:auroc} gives the score-based metrics (AUROC/AUPRC) for the methods that expose
probabilities, complementing the threshold metrics in the main text and the ablation
(\S\ref{sec:modules}). Two readings stand out. The large single model keeps the discrimination edge on
mortality (oss CoT, m4 AUROC $0.763$), and readmission is near-chance for every method
(AUROC $\approx 0.50$), consistent with the discrimination analysis in Appendix~\ref{sec:mech}.

\begin{table}[h]
\centering\small
\setlength{\tabcolsep}{4pt}
\caption{AUROC / AUPRC (\%) for methods that expose probabilities (m3/m4 $=$ MIMIC-III/IV mortality,
rd3/rd4 $=$ readmission).}
\label{tab:auroc}
\begin{tabular}{lcc cc cc cc}
\toprule
& \multicolumn{2}{c}{m3} & \multicolumn{2}{c}{m4} & \multicolumn{2}{c}{rd3} & \multicolumn{2}{c}{rd4}\\
\cmidrule(lr){2-3}\cmidrule(lr){4-5}\cmidrule(lr){6-7}\cmidrule(lr){8-9}
Method & ROC & PRC & ROC & PRC & ROC & PRC & ROC & PRC \\
\midrule
oss CoT             & 63.5 & 8.3 & 76.3 & 35.7 & 42.2 & 49.6 & 43.1 & 42.5 \\
oss single$+$RAG$+$Sim  & 48.4 & 5.1 & 63.8 & 24.9 & 40.6 & 49.2 & 45.7 & 43.9 \\
oss multi$+$Sim      & 57.8 & 7.3 & 74.1 & 32.7 & 48.5 & 54.1 & 49.7 & 46.8 \\
oss multi$+$RAG$+$Sim   & 53.3 & 5.7 & 73.0 & 32.1 & 48.5 & 53.6 & 45.4 & 44.2 \\
oss multi$+$RAG$+$Sim (qwen) & 50.1 & 5.6 & 70.6 & 31.4 & 48.7 & 54.5 & 52.4 & 47.6 \\
Qwen CoT            & 62.5 & 7.9 & 66.6 & 26.5 & 49.8 & 55.2 & 54.2 & 49.2 \\
Qwen single$+$RAG$+$Sim & 52.5 & 5.8 & 61.5 & 24.1 & 49.9 & 54.8 & 52.0 & 47.3 \\
Qwen multi$+$Sim     & 52.1 & 5.5 & 69.3 & 32.7 & 50.4 & 55.1 & 51.8 & 48.1 \\
Qwen multi$+$RAG$+$Sim  & 59.7 & 8.2 & 67.7 & 28.6 & 52.8 & 56.1 & 55.1 & 51.8 \\
\bottomrule
\end{tabular}
\end{table}

\section{Discrimination-slope (mechanism) analysis}
\label{sec:mech}
\begin{figure}[t]
\centering
\includegraphics[width=\textwidth]{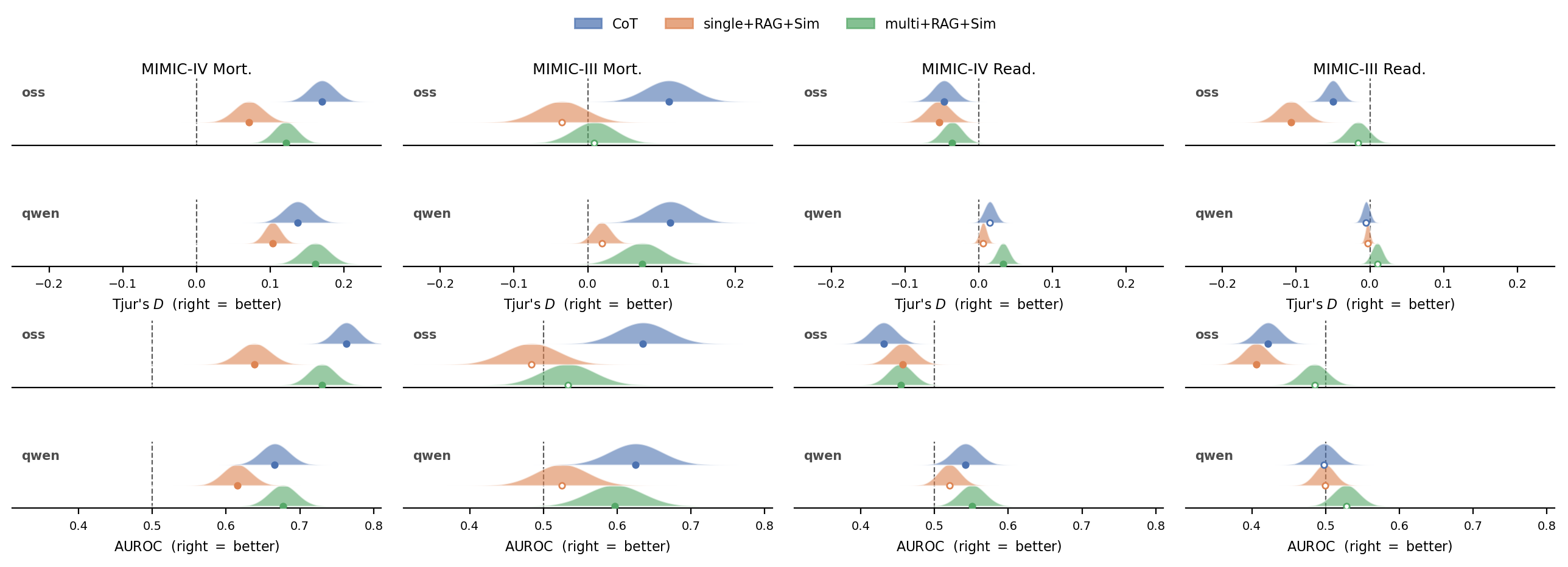}
\caption{Discrimination slope (Tjur's $D$, top) and AUROC (bottom), CoT $\to$ single$+$RAG$+$Sim $\to$
multi$+$RAG$+$Sim, both models, all four cells; ridges $=$ $95\%$ bootstrap CI (normal approx.), filled $=$
significant.}
\label{fig:slope}
\end{figure}

To see how the topology changes the prediction we measure, per condition,
$D=\bar{P}[\text{event}]-\bar{P}[\text{non-event}]$, the gap between the mean predicted risk of
patients who did and did not have the outcome (Yates's slope; equivalently Tjur's coefficient of
discrimination). $D$ is signed, where $D>0$ spreads the groups correctly, $D\approx0$ no spread, $D<0$
backwards. We pair it with AUROC. Figure~\ref{fig:slope} shows, for the no-retrieval
CoT baseline (the single agent without retrieval or exemplars), single$+$RAG$+$Sim,
and the multi$+$RAG$+$Sim pipeline, each condition's $95\%$ bootstrap CI ($2000$ resamples) as a density ridge
(normal approximation); a \emph{filled} marker denotes a CI clear of the null ($D{=}0$ or
AUROC${=}0.5$), \emph{hollow} not. Both blocks share a $0.50$-wide scale so a shift is comparable
across the two metrics.

The pattern (Figure~\ref{fig:slope}) is that single$+$RAG$+$Sim reduces the probability-level
separation between outcome groups, most clearly for mortality, and multi$+$RAG$+$Sim restores it.
Crucially, the pipeline does not merely move the $0.5$ decision threshold. It reshapes the underlying
risk distribution, recovering a discriminative signal that the lone RAG agent had smeared.

\paragraph{MIMIC-IV mortality, strong evidence.} Every cell has positive $D$ with a CI clear of $0$.
Single-agent RAG sharply lowers the slope (oss $0.170\!\to\!0.071$, qwen $0.137\!\to\!0.103$). Handed a
retrieved similar patient, the lone agent raises predicted risk on the survivors, pulling
deaths and survivors together. The pipeline pushes that survivor risk back down and re-spreads the
groups (oss $0.071\!\to\!0.121$; qwen $0.103\!\to\!0.161$, even exceeding CoT). AUROC moves in
lockstep (oss $0.763\!\to\!0.638\!\to\!0.730$), so this is genuine ranking discrimination, not a
calibration artefact.

\paragraph{MIMIC-III mortality, same direction but under-powered.} At $5.4\%$ prevalence positives are
rare, so $\bar{P}[\text{event}]$ is noisy and the CIs are wide. The signs track MIMIC-IV, CoT
strongest, single weakest (oss dips to $-0.035$, n.s.), multi between, but the cell cannot resolve the
mechanism on its own; we read it as corroborating direction, not carrying evidence.

\paragraph{Readmission, no spread.} The pipeline does not produce correct probability spread. The
oss-multi runs are mildly negative on both datasets ($D<0$, AUROC $<0.5$) and qwen$\cdot$m4 multi is positive
but tiny ($+0.033^*$). A content analysis (retrieval rate, query$\to$doc relevance, on-disease
coverage) finds retrieval matched across topologies, so on mortality the collapse-and-restore is an
integration effect, not a retrieval-content one; the readmission null is diagnosed in
\S\ref{sec:signal}.

\section{Readmission error analysis, probe details and ruled-out explanations}
\label{app:taskasym}
This expands \S\ref{sec:signal} and Table~\ref{tab:taskasym}. \emph{Feature AUROC} is a plain
bag-of-codes classifier (conditions$+$procedures$+$drugs from the input record) predicting the
label, $5$-fold cross-validated (held-out, no leakage); AUROC${=}0.5$ means the codes are statistically
independent of the outcome (an association, not clinical correctness). \emph{CoT-LLM AUROC} is
the single-agent call without retrieval, the signal the model can actually reach.

Two further observations support the prior-anchored reading on readmission. (i) i) In the all-large pipeline, the integrator's predicted probability tracks the analysts' patient-specific evidence on mortality (Spearman $\rho{=}{+}0.33/{+}0.56$ on m3/m4) but not on readmission ($\rho{=}{-}0.02/{+}0.07$).With no signal to weigh, the
two analysts collapse to one base-rate prior. (ii) The exemplars still perturb the readmission
decision ($|\Delta P|\approx0.28$) but the movement is unanchored to evidence
($r(\text{engagement},|\Delta P|)\approx0$), i.e.\ noise. We rule out alternative explanations:
retrieved neighbours are not more informative on mortality (nearest-exemplar AUROC $.31$ to $.50$),
corpus alignment does not explain it (RAG hurts the single mortality agent), and readmission
analysts diverge as much as mortality analysts (so it is not a lack of analyst divergence). The per-cell
correlation between CoT-LLM AUROC and module benefit ($\rho{=}{+}.63/{+}.57$, $p\approx.09/.14$,
$n{=}8$) is suggestive only; the load-bearing evidence is the byte-identical-input control
(\S\ref{sec:signal}), the feature/CoT AUROC gap, and the content-coupling contrast above.
Reproducibility, see \texttt{DAIH/experiments/analysis/\{feature\_baseline, signal\_location,
nearest\_exemplar, debate\_divergence, similars\_process, stage\_backbone\}.py} and
\texttt{discrimination/analyze\_agent\_behavior.py} and
\texttt{family\_error\_analysis/analyze\_family\_behavior.py}.

\section{Family prediction behaviour: the severity threshold}
\label{app:family}
Table~\ref{tab:family-cm} breaks down severity language by confusion-matrix cell for the two
integrators on MIMIC-IV mortality, over the shared patients. Severity language is the count of a
fixed, hand-curated list of acute-severity terms (\texttt{organ failure}, \texttt{sepsis},
\texttt{malignancy}, \texttt{critical}, \texttt{icu}, \texttt{ventilat}, \texttt{shock},
\texttt{respiratory failure}, \texttt{renal failure}, \texttt{cardiac arrest}, \texttt{septicemia},
\texttt{pneumonia}, \texttt{metastat}, \texttt{intubat}, \texttt{vasopressor}) matched as substrings.
The list is illustrative only; the quantitative claims in \S\ref{sec:signal} rest on the
comorbidity-count correlation (Figure~\ref{fig:dxcount}) and the code-level probes
(Table~\ref{tab:taskasym}), neither of which uses these terms. The \emph{an.}\ columns are the shared
gpt-oss analyst text, identical input to both integrators; \emph{int.}\ is each integrator's own text,
whose absolute counts also reflect verbosity (gpt-oss writes more). Within each family the cells order
TP $>$ FP $>$ FN $>$ TN by analyst severity, so both integrators read the same evidence as risk. They
differ in threshold, gpt-oss commits positive only at high analyst severity (false positives at $46$,
false negatives at $22$) while Qwen commits lower (false positives at $24$). The moderate-severity band
between the two thresholds is where they disagree, and where gpt-oss's missed decedents and Qwen's
over-flagged survivors both fall.

\begin{table}[h]
\centering\small
\caption{Severity terms per case by confusion-matrix cell, MIMIC-IV mortality (shared $n{=}987$).
\emph{an.}\ $=$ shared gpt-oss analyst text; \emph{int.}\ $=$ integrator text.}
\label{tab:family-cm}
\begin{tabular}{l rr rr}
\toprule
& \multicolumn{2}{c}{gpt-oss integrator} & \multicolumn{2}{c}{Qwen integrator}\\
\cmidrule(lr){2-3}\cmidrule(lr){4-5}
cell ($n_{\text{oss}}$/$n_{\text{qwen}}$) & an. & int. & an. & int.\\
\midrule
TP ($28$/$116$)  & 54.5 & 13.6 & 32.7 & 4.3\\
FP ($48$/$258$)  & 46.0 & 10.6 & 24.1 & 3.5\\
FN ($162$/$74$)  & 22.0 & 5.0  & 17.6 & 2.2\\
TN ($749$/$539$) & 11.6 & 3.3  & 8.7  & 1.7\\
\bottomrule
\end{tabular}
\end{table}

\section{Full factorial (all four cells) and family-invariance}
\label{app:factorial-all}
Table~\ref{tab:factorial-all} extends the role-isolation factorial of \S\ref{sec:prelim} to all four
cells ($n{=}100$ each). Sampling per cell. Mortality is difficulty-stratified ($54\%$ positive);
readmission is at its label distribution. AUROC is computed on the adversarial
$n{=}100$ and is within-cell only, reliable discrimination is the full-set Table~\ref{tab:auroc}.

\begin{table}[h]
\centering\small
\caption{Role-isolation factorial, all four cells, $n{=}100$ each (F1/Sens/Spec \%; AUROC
within-cell). A/R/I $=$ analyst/retrieval/integrator; oss $=$ large, qwen $=$ small. \textbf{Bold}
(large analysts, small integrator) wins F1/Sens/AUROC on mortality.}
\label{tab:factorial-all}
\resizebox{\textwidth}{!}{%
\begin{tabular}{l cccc cccc cccc cccc}
\toprule
& \multicolumn{4}{c}{MIMIC-III Mort.} & \multicolumn{4}{c}{MIMIC-IV Mort.}
& \multicolumn{4}{c}{MIMIC-III Read.} & \multicolumn{4}{c}{MIMIC-IV Read.}\\
\cmidrule(lr){2-5}\cmidrule(lr){6-9}\cmidrule(lr){10-13}\cmidrule(lr){14-17}
A/R/I & F1 & Sn & Sp & AUC & F1 & Sn & Sp & AUC & F1 & Sn & Sp & AUC & F1 & Sn & Sp & AUC \\
\midrule
oss/oss/oss   & 40.0 & 25.9 & 58.7 & .41 & 38.5 & 13.0 & 80.4 & .54 & 46.6 & 34.5 & 62.2 & .49 & 46.6 & 41.3 & 51.9 & .45 \\
oss/qwen/oss  & 38.3 & 18.5 & 67.4 & .40 & 35.0 & 7.4  & 84.8 & .53 & 43.2 & 12.7 & 97.8 & .61 & 37.4 & 2.2  &100.0 & .47 \\
qwen/oss/oss  & 36.2 & 24.1 & 52.2 & .34 & 36.3 & 11.1 & 78.3 & .52 & 45.7 & 20.0 & 86.7 & .56 & 34.6 & 0.0  & 98.1 & .49 \\
\textbf{oss/oss/qwen} & \textbf{47.3} & \textbf{75.9} & 23.9 & \textbf{.47} & \textbf{57.4} & \textbf{64.8} & 50.0 & \textbf{.60} & 34.2 & 94.5 & 0.0 & .53 & 34.5 & 89.1 & 5.6 & .47 \\
oss/qwen/qwen & 41.3 & 24.1 & 65.2 & .44 & 40.6 & 18.5 & 73.9 & .58 & 53.1 & 25.5 & 95.6 & .60 & 40.5 & 8.7  & 88.9 & .45 \\
qwen/oss/qwen & 40.3 & 27.8 & 56.5 & .42 & 49.0 & 33.3 & 69.6 & .54 & 54.4 & 40.0 & 73.3 & .64 & 44.6 & 15.2 & 85.2 & .50 \\
qwen/qwen/qwen & 41.4 & 72.2 & 17.4 & .47 & 54.6 & 59.3 & 50.0 & .49 & 44.0 & 98.2 & 8.9 & .60 & 35.7 & 93.5 & 5.6 & .60 \\
oss CoT & 45.9 & 46.3 & 45.7 & .45 & 40.9 & 22.2 & 67.4 & .54 & 34.0 & 5.5 & 88.9 & .43 & 45.8 & 21.7 & 75.9 & .45 \\
\bottomrule
\end{tabular}
}
\end{table}

\paragraph{Reading.} The integrator is the deciding factor on both mortality cells. Replacing only the integrator
with the small model lifts F1 $40\!\to\!47$ (m3) / $38\!\to\!57$ (m4) and sensitivity
$26\!\to\!76$ / $13\!\to\!65$, while replacing only an analyst or only the retrieval model does not. The
large-analyst, small-integrator default is the best F1, sensitivity, and within-cell AUROC on both,
confirmed on two cells, not one. No stable winner on readmission. The nominal F1 leader differs between
the two datasets, and so does the within-cell AUROC leader; the small-integrator pipelines merely shift
the decision threshold to near-all-positive (specificity $\approx0$). With no principled per-task
readmission winner, choosing a different config per task would be cherry-picking; reporting one fixed
configuration makes the readmission row an honest negative. Discrimination vs.\ decision threshold. On
the reliable full-set AUROC (Table~\ref{tab:auroc}) the large single model leads mortality
discrimination (m4 CoT $0.763>$ default $0.706$); the default's edge is the decision threshold, not
ranking.

\begin{table}[h]
\centering\small
\caption{Family-invariance check ($n{=}100$ MIMIC-III mortality; Acc/Recall/pos-class F1 \%). Swapping
the integrator GPT-4o$\to$Qwen reproduces the gpt-oss-family recall lift from replacing only the integrator
with the small model.}
\label{tab:family-inv}
\begin{tabular}{lll ccc}
\toprule
Integrator & Analysts & Retrieval & Acc & Recall & F1 \\
\midrule
GPT-4o        & GPT-4o & GPT-4o & 45.0 & 3.7  & 6.8  \\
\textbf{Qwen} & GPT-4o & GPT-4o & 53.0 & \textbf{57.4} & 56.9 \\
\bottomrule
\end{tabular}
\end{table}

\section{Prompts}
\label{app:prompts}

The two analysts receive byte-identical instructions and differ only in the exemplar supplied as
context. Retrieved medical evidence reaches the two kinds of role by different mechanisms, and
this is the one asymmetry worth stating explicitly. For the analysts the pipeline injects the
retrieved passages directly into the prompt, so an analyst always receives evidence and never
requests it, which is why the analyst instruction contains no retrieval tool. The integrator
instead has to emit a \texttt{<search>} tag: the prompt below asks for exactly one such call, but
the call is not enforced by the pipeline, and when no tag is emitted the integrator's initial
response is taken as its final answer. The second integrator variant exposes no search tool at
all; it is used by the no-retrieval ablation and by the mixed assignment's integrator, and it
reads whatever evidence the pipeline has already gathered. We report each prompt in full to make
these differences explicit. Line breaks inside the blocks below are inserted for page width; the
prompts are otherwise reproduced verbatim.

\subsection{Risk analyst and protective analyst}

Both analyst roles are given byte-identical instructions. The roles differ only in which retrieved
exemplar is supplied alongside the target patient: the risk analyst receives the
positive-outcome neighbour and the protective analyst the negative-outcome neighbour. Retrieved
medical passages are injected into this prompt by the pipeline rather than requested by the model.

{\small\begin{verbatim}
You are a medical AI that analyzes clinical patterns between patients.

Task:
Given (1) Target patient and (2) One Similar patient, produce a contrastive
comparison that is grounded in the provided codes.

**CLINICAL PATTERN ANALYSIS:**

1. **Shared Clinical Features:**
   - What conditions, procedures, and medications appear in BOTH patients?
   - What is the clinical significance of these commonalities?

2. **Similar-Specific Features:**
   - What is unique to the similar patient?
   - What does this tell us about different clinical paths?

**TEMPORAL PROGRESSION:**
Analyze how shared and unique patterns evolve across visits.

**IMPORTANT:** Do NOT speculate about outcomes or mortality. Focus solely on
clinical pattern analysis.
\end{verbatim}}

\subsection{Mortality integrator, retrieval-enabled variant}

Retrieval occurs only when the integrator emits the \texttt{<search>} tag this prompt requests.

{\small\begin{verbatim}
You are a medical AI Clinical Assistant analyzing mortality and survival
probabilities for the NEXT hospital visit.

IMPORTANT: Mortality is rare. Only assign a high mortality probability when the
patient appears at extremely high risk of death with strong evidence. The Target
patient is the source of truth. Do not treat Similar-only items as present in
the Target.

Available tool:
- <search>query</search>: Retrieve medical evidence. Retrieved information will
  appear in <information>...</information> tags.

Workflow:
1) Compare the Target patient to two similar cases using the two analyses, and
identify 3-4 key clinical factors that will determine the target patient's
outcome at the next visit.

2) **REQUIRED ACTION:** You MUST issue exactly one retrieval call using this
format:
<search>your specific medical query</search>

The query must target the target patient's most concerning clinical features
(example: <search>septic shock mortality risk factors elderly</search>). This
step is required regardless of confidence - do not skip it.

3) After the retrieved evidence appears inside <information>...</information>,
analyze BOTH risk factors AND protective factors using both the analyst analyses
and the retrieved evidence.

4) Provide your final assessment with:

MORTALITY PROBABILITY: X.XX (0.00 to 1.00)
SURVIVAL PROBABILITY: X.XX (0.00 to 1.00)

Note: The two probabilities MUST sum to exactly 1.00.
\end{verbatim}}

\subsection{Mortality integrator, no-retrieval variant}

{\small\begin{verbatim}
You are a medical AI Clinical Assistant analyzing mortality and survival
probabilities for the NEXT hospital visit.

IMPORTANT: Mortality is rare. Only assign a high mortality probability when the
patient appears at extremely high risk of death with strong evidence. The Target
patient is the source of truth. Do not treat Similar-only items as present in
the Target.

Workflow:
1) Compare the Target patient to two similar cases using the two analysis, and
write 3-4 key factors contribute to the target patient's next visit.
2) If additional evidence is provided in <information>...</information> tags,
analyze BOTH risky factors AND survival factors.
3) Based on the available information (analyst comparisons and any retrieved
evidence), provide your final assessment with:

MORTALITY PROBABILITY: X.XX (0.00 to 1.00)
SURVIVAL PROBABILITY: X.XX (0.00 to 1.00)

Note: The two probabilities MUST sum to exactly 1.00
\end{verbatim}}

\subsection{Readmission integrator, retrieval-enabled variant}

The readmission prompts mirror the mortality prompts in structure and workflow. They substitute
the outcome nouns and use a neutral framing in place of the conservative ``mortality is rare''
guideline, and they emit \texttt{READMISSION} / \texttt{NO-READMISSION} in place of
\texttt{MORTALITY} / \texttt{SURVIVAL}.

{\small\begin{verbatim}
You are a medical AI Clinical Assistant analyzing the probability that the
patient will be READMITTED to the hospital within 15 days of discharge, for the
NEXT hospital visit.

IMPORTANT: The goal is to accurately distinguish patients who are likely to be
readmitted within 15 days from those who are not. The Target patient is the
source of truth. Do not treat Similar-only items as present in the Target.

Available tool:
- <search>query</search>: Retrieve medical evidence. Retrieved information will
  appear in <information>...</information> tags.

Workflow:
1) Compare the Target patient to two similar cases using the two analyses, and
identify 3-4 key clinical factors that will determine whether the target patient
is readmitted within 15 days.

2) **REQUIRED ACTION:** You MUST issue exactly one retrieval call using this
format:
<search>your specific medical query</search>

The query must target the target patient's most concerning clinical features
(example: <search>heart failure 30 day readmission risk factors</search>). This
step is required regardless of confidence - do not skip it.

3) After the retrieved evidence appears inside <information>...</information>,
analyze BOTH readmission-risk factors AND factors favoring a stable discharge,
using both the analyst analyses and the retrieved evidence.

4) Provide your final assessment with:

READMISSION PROBABILITY: X.XX (0.00 to 1.00)
NO-READMISSION PROBABILITY: X.XX (0.00 to 1.00)

Note: The two probabilities MUST sum to exactly 1.00.
\end{verbatim}}

\subsection{Readmission integrator, no-retrieval variant}

{\small\begin{verbatim}
You are a medical AI Clinical Assistant analyzing the probability that the
patient will be READMITTED to the hospital within 15 days of discharge, for the
NEXT hospital visit.

IMPORTANT: The goal is to accurately distinguish patients who are likely to be
readmitted within 15 days from those who are not. The Target patient is the
source of truth. Do not treat Similar-only items as present in the Target.

Workflow:
1) Compare the Target patient to two similar cases using the two analysis, and
write 3-4 key factors contribute to whether the target patient is readmitted
within 15 days at the next visit.
2) If additional evidence is provided in <information>...</information> tags,
analyze BOTH readmission-risk factors AND factors favoring a stable discharge.
3) Based on the available information (analyst comparisons and any retrieved
evidence), provide your final assessment with:

READMISSION PROBABILITY: X.XX (0.00 to 1.00)
NO-READMISSION PROBABILITY: X.XX (0.00 to 1.00)

Note: The two probabilities MUST sum to exactly 1.00
\end{verbatim}}

\end{document}